\documentclass{article} 
\usepackage{iclr2022_conference,times}

\usepackage{amsmath,amsfonts,bm}

\def\eqref#1{equation~\ref{#1}}

\def\1{\bm{1}}

\DeclareMathAlphabet{\mathsfit}{\encodingdefault}{\sfdefault}{m}{sl}
\SetMathAlphabet{\mathsfit}{bold}{\encodingdefault}{\sfdefault}{bx}{n}

\newcommand{\E}{\mathbb{E}}

\newcommand{\R}{\mathbb{R}}

\DeclareMathOperator*{\argmax}{arg\,max}
\DeclareMathOperator*{\argmin}{arg\,min}

\usepackage{hyperref}
\usepackage{url}
\usepackage{soul}
\usepackage{amsmath,amsfonts,amsthm,amssymb,amsopn,bm,color,mathtools,booktabs,multirow,appendix,caption,hyperref,wrapfig,algorithm,algorithmic}

\usepackage{enumitem}
\setlist[itemize]{noitemsep, topsep=0pt, leftmargin=11pt}
\def\E{\mathbb{E}}
\def\P{\mathbb{P}}
\def\R{\mathbb{R}}
\def\1{\bm{1}}
\newcommand{\mb}[1]{\mathbf{#1}}
\newcommand{\mc}[1]{\mathcal{#1}}

\usepackage{xr}
\makeatletter
\newcommand*{\addFileDependency}[1]{
  \typeout{(#1)}
  \@addtofilelist{#1}
  \IfFileExists{#1}{}{\typeout{No file #1.}}
}
\makeatother

\newcommand*{\myexternaldocument}[1]{
    \externaldocument{#1}
    \addFileDependency{iclr/#1.tex}
    \addFileDependency{./#1.aux}
}
\myexternaldocument{appendix}

\title{Learning to Actively Learn: A Robust Approach}

\author{Jifan Zhang \\
Department of Computer Science\\
University of Wisconsin\\
Madison, WI, USA \\
\texttt{jifan@cs.wisc.edu} \\
\And
Lalit Jain \\
Foster School of Business \\
University of Washington \\
Seattle, WA, USA \\
\texttt{lalitj@uw.edu} \\
\And
Kevin Jamieson \\
Allen School of Computer Science \& Engineering \\
University of Washington \\
Seattle, WA, USA \\
\texttt{jamieson@cs.washington.edu}
}

\iclrfinalcopy 
\begin{document}

\maketitle

\begin{abstract}
This work proposes a procedure for designing algorithms for specific adaptive data collection tasks like active learning and pure-exploration multi-armed bandits. 
Unlike the design of traditional adaptive algorithms that rely on concentration of measure and careful analysis to justify the correctness and sample complexity of the procedure, our adaptive algorithm is learned via adversarial training over equivalence classes of problems derived from information theoretic lower bounds. 
In particular, a single adaptive learning algorithm is learned that competes with the best adaptive algorithm learned for each equivalence class.
Our procedure takes as input just the available queries, set of hypotheses, loss function, and total query budget.
This is in contrast to existing meta-learning work that learns an adaptive algorithm relative to an explicit, user-defined subset or prior distribution over problems which can be challenging to define and be mismatched to the instance encountered at test time.
This work is particularly focused on the regime when the total query budget is very small, such as a few dozen, which is much smaller than those budgets typically considered by theoretically derived algorithms. 
We perform synthetic experiments to justify the stability and effectiveness of the training procedure, and then evaluate the method on tasks derived from real data including a noisy 20 Questions game and a joke recommendation task.
\end{abstract}

\section{Introduction} \label{sec:intro}

Closed-loop learning algorithms use previous observations to inform what measurements to take next in a closed loop in order to accomplish inference tasks far faster than any fixed measurement plan set in advance.
For example, active learning algorithms for binary classification have been proposed that under favorable conditions require exponentially fewer labels than passive, random sampling to identify the optimal classifier \citep{hanneke2014theory,katz2021improved}.
And in the multi-armed bandits literature, adaptive sampling techniques have demonstrated the ability to identify the ``best arm'' that optimizes some metric 
with far fewer experiments than a fixed design \citep{garivier2016optimal,fiez2019sequential}.  
Unfortunately, such guarantees often either require simplifying assumptions that limit robustness and applicability, or algorithmic use of concentration inequalities that are very loose unless the number of samples is very large.

This work proposes a framework for producing algorithms that are learned through simulated experience to be as effective and robust as possible, even on a tiny measurement budget (e.g., 20 queries) where most theoretical guarantees do not apply.
Our work fits into a recent trend sometimes referred to as \emph{learning to actively learn} and \emph{differentiable meta-learning in bandits} \citep{konyushkova2017learning, bachman2017learning,fang2017learning,boutilier2020differentiable,kveton2020differentiable} which tune existing algorithms or learn entirely new active learning algorithms by policy optimization. 
Previous works in this area learn a policy by optimizing with respect to data observed through prior experience (e.g., meta-learning or transfer learning) or an assumed explicit prior distribution of problem parameters (e.g. a Gaussian prior over the true weight vector for linear regression).
In contrast, our approach makes no assumptions about what parameters are likely to be encountered at test time, and therefore produces algorithms that \emph{do not suffer from mismatching priors at test time}.
Instead, our method learns a policy that attempts to mirror the guarantees of frequentist algorithms with instance dependent sample complexities: there is an intrinsic difficulty measure that orders problem instances and given a fixed budget, higher accuracies can be obtained for \emph{all} easier instances than harder instances.  
This difficulty measure is most naturally derived from information theoretic lower bounds. 


%

But unlike information theoretic bounds that hand-craft adversarial instances, inspired by the robust reinforcement learning literature, we formulate a novel adversarial training objective that automatically train minimax policies and propose a tractable and computationally efficient relaxation.
This allows our learned policies to be very aggressive while maintaining robustness over difficulty in problem instances, without resorting to using loose concentration inequalities in the algorithm.
Indeed, this work is particularly useful in the setting where relatively few rounds of querying can be made. 
The learning framework is general enough to be applied to many active learning settings of interest and is intended to be used to produce robust and high performing algorithms.
We implement the framework for the pure-exploration combinatorial bandit problem —
a paradigm including problems such as active binary classification and the 20 question game.
We empirically validate our framework on a simple synthetic experiment before turning our attention to datasets derived from real data including a noisy 20 Questions game and a joke recommendation task which are also embedded as combinatorial bandits. 
As demonstrated in our experiments, in the low budget setting, our learned algorithms are the only ones that \emph{both enjoy robustness guarantees} (as opposed to greedy and existing learning to actively learn methods) \emph{and perform non-vacuously and instance-optimally} (as opposed to statistically justified algorithms).

\section{Proposed Framework for Robust Learning to Actively Learn} \label{sec:framework}

From a birds-eye perspective, whether learned or defined by an expert, any algorithm for active learning can be thought of as a policy from the perspective of reinforcement learning. 
To be precise, at time $t$, based on an internal state $s_t$, the policy $\pi$ defines a distribution $\pi(s_t)$ over the set of potential actions $\mc{X}$. It then takes action $x_t\in \mc{X}, x_t\sim  \pi(s_t)$ and receives observation $y_t$, updates the state and the process repeats.

Fix a horizon $T \in \mathbb{N}$, and a problem instance $\theta_* \in \Theta \subseteq \R^d$ which parameterizes the observation distribution. For $t=1,2,\dots,T$ 
\begin{itemize}
\item state $s_t \in \mc{S}$ is a function of the history, $\{(x_i,y_i)\}_{i=1}^{t-1}$,
\item action $x_t \in \mc{X}$ is drawn at random from the distribution $\pi(s_t)$ defined over $\mc{X}$, and
\item next state $s_{t+1} \in \mc{S}$ is constructed by taking action $x_t$ in state $s_t$ and observing $y_t \sim f( \cdot | \theta_*, s_t, x_t)$
\end{itemize}
until the game terminates at time $t=T$ and the learner receives a loss $L_T$ which is task specific. Note that $L_T$ is a random variable that depends on the tuple $(\pi,\{(x_i,y_i)\}_{i=1}^T,\theta_*)$.
We assume that $f$ is a known parametric distribution to the policy but the parameter $\theta$ is unknown to the policy. 
Let $\P_{\pi,\theta}, \E_{\pi,\theta}$ denote the probability and expectation under the probability law induced by executing policy $\pi$ in the game with $\theta_*=\theta$ to completion.
Note that $\P_{\pi,\theta}$ includes any internal randomness of the policy $\pi$ and the random observations $y_t \sim f(\cdot|\theta,s_t,x_t)$. Thus, $\P_{\pi,\theta}$ assigns a probability to any trajectory $\{(x_i,y_i)\}_{i=1}^T$.
For a given policy $\pi$ and $\theta_*=\theta$, the metric of interest we wish to minimize is the expected loss 
$\ell(\pi,\theta) := \E_{\pi,\theta}\left[ L_T \right]$
where $L_T$ as defined above is the loss observed at the end of the episode.
For a fixed policy $\pi$, $\ell(\pi,\theta)$ defines a loss surface over all possible values of $\theta$.
This loss surface captures the fact that some values of $\theta$ are just intrinsically harder than others, but also that a policy may be better suited for some values of $\theta$ versus others.

Finally, we assume we are equipped with a positive function $\mc{C}: \Theta \rightarrow (0,\infty)$ that assigns a score to each $\theta \in \Theta$ that intuitively captures the ``difficulty'' of a particular $\theta$, and can be used as a partial ordering of $\Theta$.
Ideally, $\mc{C}(\theta)$ is a monotonic transformation of $\ell(\pi^*,\theta)$ for some ``best'' policy $\pi^*$ that we will define shortly.
Our plan is now as follows, in Section~\ref{sec:linear_bandit}, we ground the discussion and describe $\mc{C}(\theta)$ for the combinatorial bandit problem. Then in Section~\ref{sec:problem_definition}, we zoom out to define our main objective of finding a \emph{min-gap} optimal policy, finally providing an adversarial training approach in Section~\ref{sec:training_algorithm}.   

\subsection{Complexity for Combinatorial Bandits} \label{sec:linear_bandit}
A concrete example of the framework above is the combinatorial bandit problem.  
The learner has access to sets $\mc{X} = \{e_1, \cdots, e_d\}\subset \mathbb{R}^d$, where $e_i$ is the $i$-th standard basis vector, and $\mc{Z}\subset\{0,1\}^d$. In each round the learner chooses an $x_t\in \mc{X}$ according to a policy $\pi(\{(x_i,y_i)\}_{i=1}^{t-1})$ and observes $y_t$ with $\E[y_t|x_t, \theta_{\ast}] = \langle x_t, \theta_{\ast}\rangle$ for some unknown $\theta_{\ast}\in \mathbb{R}^d$. The goal of the learner is \textsc{Best Arm Identification}.
Denote $z_{\ast}(\theta_{\ast}) = \argmax_{z\in \mc{Z}} \langle z, \theta_{\ast}\rangle$, then at time $T$ the learner outputs a recommendation $\hat{z}$ and incurs loss  $L_{BAI,T} = \1\{z_{\ast}\neq \hat{z}\}$. 
This setting naturally captures the 20 question game. Indeed assume there are $d\gg T=20$ potential yes/no questions that can be asked at each time, corresponding to the elements of $\mc{X}$, and that each element of $\mc{Z}$ is a binary vector representing the answers to these questions for a given item. 
If answers $y_t$ are deterministic then $\theta_* \in \{-1,1\}^d$, but this framework also captures the case $\theta_* \in [-1,1]^d$ when answers are stochastic, or answered incorrectly with some probability.   
Then a policy $\pi$ at each time decides which question to ask based on the answers so far to determine the item closest to an unknown vector $\theta_*$.


As described in Sections~\ref{sec:related_work} and Appendix~\ref{sec:instance_dep_applications}, combinatorial bandits generalizes standard multi-armed bandits, and all of binary classification, and thus has received a tremendous amount of interest in recent years. 
A large portion of this work has focused on providing precise characterization of the information theoretic limit on the mimimal number of samples needed to identify $z_{\ast}(\theta_{\ast})$ with high probability a quantity denoted as $\rho_{\ast}(\theta_{\ast})$ which is the solution to an optimization problem \citep{soare2014best, fiez2019sequential, degenne2020gamification}
$\rho_{\ast}(\theta_\ast)^{-1} := \max_{\lambda\in \triangle_\mc{X}}\min_{\substack{\theta'\in \Theta\\  z_{\ast}(\theta') \neq z_{\ast}(\theta_{\ast})}} \sum_{x\in \mc{X}} \lambda_{x}\langle x,\theta_{\ast} - \theta'\rangle^2$
for some set of alternatives $\Theta$. This quantity provides a natural complexity measure $\mc{C}(\theta_{\ast}) = \rho_{\ast}(\theta_{\ast})$ for a given instance $\theta_{\ast}$ and we describe it in a few specific cases below.


As a warmup example, consider the standard best-arm identification problem where $\mc{Z} = \mc{X} = \{ \mb{e}_i : i \in [d]\}$ and choosing action $x_t\in \mc{X}$ results in reward $y_t \sim \text{Bernoulli}(\theta_{i_t})$. Let $i_\ast(\theta) = \argmax_{z\in \mc{Z}} z^{\top}\theta =  \argmax_i \theta_i$. Then in this case $\rho_{\ast}(\theta) \approx \sum_{i \neq i_\ast(\theta)} (\theta_{i_\ast(\theta)} - \theta_i)^{-2}$
and it's been shown that there exists a constant $c_0>0$ such that for any sufficiently large $\nu >0$ we have
\begin{align*}
    \min_\pi \max_{\theta : \rho_{\ast}(\theta) \leq \nu}\ell_{BAI}(\pi,\theta) \geq \exp( - c_0 T / \nu)
\end{align*}
In other words, more difficult instances correspond to $\theta$ with a small gap between the best arm and any other arm.
Moreover, for any $\theta \in \R^d$ there exists a policy $\widetilde{\pi}$ that achieves $\ell(\widetilde{\pi},\theta) \leq c_1 \exp( - c_2 T / \rho_{\ast}(\theta_{\ast}))$ where $c_1,c_2$ capture constant and low-order terms \citep{carpentier2016tight,karnin2013almost,garivier2016optimal}.
Said plainly, the above correspondence between the lower bound and the upper bound for the multi-armed bandit problem shows that $\rho_{\ast}(\theta_{\ast})$ 
is a natural choice for $\mc{C}(\theta)$ in this setting.

In recent years, algorithms for the more general combinatorial bandit setting have been established with instance-dependent sample complexities matching $\rho_{\ast}(\theta_{\ast})$ (up to logarithmic factors) \citep{karnin2013almost, chen2014combinatorial, fiez2019sequential, chen2017nearly, degenne2020gamification, katz2020empirical}.
Another complexity term that appears in \cite{cao2017disagreement} for combinatorial bandits is 
\begin{align}
    \widetilde{\rho}(\theta) = \sum_{i=1}^d \max_{z : z_i \neq z_{\ast,i}(\theta)} \frac{ \|z - z_\ast(\theta)\|_2^2}{\langle z - z_\ast(\theta),\theta \rangle^2} \label{eqn:rho_hat}.
\end{align}
One can show $\rho_{\ast}(\theta) \leq \widetilde{\rho}(\theta)$ \citep{katz2020empirical} and in many cases track each other. Because $\widetilde{\rho}(\theta)$ can be computed much more efficiently compared to $\rho_\ast(\theta)$, we take $\mc{C}(\theta)=\widetilde{\rho}(\theta)$.

\subsection{Objective: Responding to All Difficulties}\label{sec:problem_definition}

As described above, though there exists algorithms for the combinatorial bandit problem that are instance-optimal in the fixed-confidence setting along with algorithms for the fixed-budget, they do not work well with small budgets as they rely on statistical guarantees. Indeed, for their guarantees to be non-vacuous, we need the budget $T$ to be sufficiently large enough to compare to the potentially large constants in upper bounds. In practice, they are so conservative that for the first 20 samples they would just sample uniformly. To overcome this, we now provide a different framework that for policy learning in a worst-case setting that is effective even in the small budget regime.

The challenge is in finding a policy that performs well across all potential problem instances simultaneously.  It is common to consider minimax optimal policies which attempt to perform well on worst case instances — but as a result, may perform poorly on easier instances. Thus, an ideal policy $\pi$ would perform uniformly well over a set of $\theta$'s that are all equivalent in ``difficulty''. 
Since each $\theta\in \Theta$ is equipped with an inherent notion of difficulty, $C(\theta)$, we can stratify the space of all possible instances by difficulty. A good policy is one whose worst case performance over all possible \emph{problem difficulties} is minimized. We formalize this idea below. 


For any set of problem instances $\widetilde{\Theta} \subset \Theta$ and $r\geq 0$ define
\begin{align*}
    \ell(\pi, \widetilde{\Theta}) :=  \max_{\theta \in \widetilde{\Theta}} \ell(\pi,\theta)\quad\quad\text{ and }\quad\quad\Theta^{(r)} := \{\theta : \mc{C}(\theta) \leq r\}.
\end{align*}
For a fixed $r >0$ (including $r=\infty$), a policy $\pi'$ that aims to minimize just $\ell(\pi',\Theta^{(r)})$ will be minimax for $\Theta^{(r)}$ and may not perform well on easy instances. 
To overcome this shortsightedness we introduce a new objective by focusing on $\ell(\pi, \Theta^{(r)}) - \min_{\pi'} \ell(\pi', \Theta^{(r)})$; the \emph{sub-optimality gap} of a given policy $\pi$ relative to an $r$-dependent baseline policy trained specifically for each $r$.
\fbox{\begin{minipage}{\textwidth}
\textbf{Objective:} Return the policy
\begin{align}
    \pi_* := \arg\min_\pi \max_{r > 0} \left( \ell(\pi, \Theta^{(r)}) - \min_{\pi'}\ell(\pi', \Theta^{(r)}) \right) \label{eqn:optimal_obj}
\end{align}
which minimizes the worst case sub-optimality gap over all $r > 0$. 
\end{minipage}}
\begin{wrapfigure}{r}{.35\textwidth}
    \begin{center}
        \includegraphics[trim={1cm 0cm 1cm 2cm},clip,width=.35\textwidth]{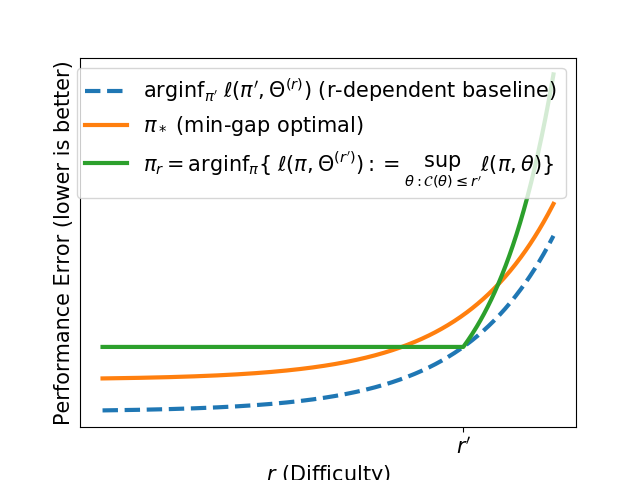}
    \end{center}
    \caption{\small Performance curves for various policies.\\}
    \label{fig:motivation}
    \vspace{-3\intextsep}
\end{wrapfigure}

Figure~\ref{fig:motivation} illustrates these definitions. The blue curve ($r$-dependent baseline) captures the best possible performance $\min_{\pi'} \ell(\pi', \Theta^{(r)})$ that is possible for each difficulty level $r$. In other words, the $r$-dependent baseline defines a different policy for each value of $r$. Therefore, the blue curve may be unachievable with just a single policy. The green curve captures a policy that achieves the minima ($r$-dependent baseline) at a given $r'$. Though it is the ideal policy for this difficulty, it could be sub-optimal at any other difficulty. The orange curve is the performance of our optimal policy $\pi^{\ast}$ — it is willing to sacrifice performance for any given $r$ to achieve an overall better worst case gap from the baseline. 


\section{MAPO: Adversarial Training Algorithm} \label{sec:training_algorithm}

Identifying $\pi_*$ naively requires the computation of $\min_{\pi'}\ell(\pi', \Theta^{(r)})$ for all $r >0$.
However, in practice 
given an increasing sequence $r_1 < \dots < r_K$ that indexes nested sets of problem instances of increasing difficulty, $\Theta^{(r_1)} \subset \Theta^{(r_2)} \subset \dots \subset \Theta^{(r_K)}$, we wish to identify a policy $\widehat{\pi}$ that minimizes the maximum sub-optimality gap with respect to this sequence. Explicitly, we seek to learn 
\begin{align}\label{eqn:opt_policy}
    \widehat{\pi} = \arg\min_\pi \max_{k \leq K}  \left( \ell(\pi, \Theta^{(r_k)}) - \ell(\pi_k, \Theta^{(r_k)}) \right) 
    \text{ where } \quad\pi_k \in \arg\min_\pi \max_{\theta : \mc{C}(\theta) \leq r_k}  \ell(\pi, \theta).
\end{align}
Note that as $K \to \infty$ and $\sup_k \frac{r_{k+1}}{r_k} \to 1$, \eqref{eqn:optimal_obj} and \eqref{eqn:opt_policy} are essentially equivalent under benign smoothness conditions on $\mc{C}(\theta)$, in which case $\widehat{\pi} \rightarrow \pi_*$.
In practice, we choose $\Theta^{(r_K)}$ contains all problems that can be solved within the budget $T$ relatively accurately, and a small $\epsilon > 0$, where $\max_k \frac{r_{k+1}}{r_k} = 1 + \epsilon$.
In Algorithm~\ref{alg:training}, our algorithm MAPO efficiently solves this objective by first computing $\pi_k$ for all $k \in [K]$ to obtain $\ell(\pi_k, \Theta^{(r_k)})$ as benchmarks, and then uses these benchmarks to train $\widehat{\pi}$. The next section will focus on the challenges of the optimization problems in \eqref{eqn:alg_obj1} and \eqref{eqn:alg_obj2}.

\begin{algorithm}
\small
\begin{algorithmic}
\STATE {\bfseries Input:}  sequence $\{r_k\}_{k=1}^K$, complexity function $\mc{C}$.
\STATE {\bfseries Define}  ${k(\theta)} \in [K]$ such that $r_{k(\theta)-1} < \mc{C}(\theta) \leq r_{k(\theta)}$ for all $\theta$ with $\mc{C}(\theta) \leq r_K$.
\FOR{$k \in 1, ..., K$}
\STATE Obtain policy $\pi_k$ by solving:
\begin{align}
    \pi_k := \argmin_\pi \ell(\pi, \Theta^{(r_k)}) =  \argmin_\pi \max_{\theta \in \Theta^{(r_k)}}\ell(\pi, \theta)\quad\quad\text{and}\quad\quad b^{(r_k)} := \ell(\pi_k, \Theta^{(r_k)})\label{eqn:alg_obj1}
\end{align}
\STATE 
\ENDFOR

\STATE \textbf{Training for min-gap optimal policy:} Solve the following:
\begin{align}
    \widehat{\pi} = \argmin_\pi \max_{\theta \in \Theta^{(r_K)}} \left[ \ell(\pi, \theta) - b^{(r_{k(\theta)})}\right] \label{eqn:alg_obj2} 
\end{align}

\noindent \textbf{Output}: $\widehat{\pi}$ (a solution to \eqref{eqn:opt_policy}).
\end{algorithmic}

\caption{MAPO: Min-gap Adversarial Policy Optimization}
\label{alg:training}
\end{algorithm}

\subsection{Differentiable policy optimization}
\label{sec:policy_opt}


The critical part of running MAPO (Algorithm~\ref{alg:training}) is to solve for \eqref{eqn:alg_obj1} and \eqref{eqn:alg_obj2}. Note that \eqref{eqn:alg_obj2} is an optimization of the same form with \eqref{eqn:alg_obj1} after shifting the loss by the scalar value $b^{(r_{k(\theta)})}$.
Consequently, to learn $\{\widetilde{\pi}_k\}_k$ and $\widehat{\pi}$, it suffices to develop a training procedure to solve $\min_\pi \max_{\theta \in \Omega} \ell'(\pi,\theta)$ for an arbitrary set $\Omega$ and generic loss function $\ell'(\pi,\theta)$.

We would like to solve this saddle-point problem using an alternating gradient descent/ascent method in Algorithm~\ref{alg:optim} that we describe now. 
Instead of optimizing over all possible policies, we restrict the policy class to neural networks that take state representation as input and output a probability distribution over actions, parameterized by weights $\psi$. 
In practice, $\ell'(\pi^\psi, \theta)$ may be poorly behaved in $(\psi,\theta)$ so a gradient  descent/ascent procedure may get stuck in a neighborhood of a critical point that is not an optimal solution to the saddle point problem.  To avoid this, we instead track over many different possible $\theta$'s (intuitively corresponding to different initializations):


\begin{align}
    \min_\psi \max_{\theta \in \Omega} \ell'(\pi^\psi,\theta) 
    &= \min_\psi \max_{\widetilde{\theta}_{1:N} \subset \Omega} \max_{i\in [N]}\ell'(\pi^\psi, \widetilde{\theta}_i). \label{eqn:max_particles}\\
    &= \min_\psi \max_{\widetilde{\theta}_{1:N} \subset \Omega} \max_{\lambda\in \Delta_{N}}\E_{i\sim \lambda}\ell'(\pi^\psi, \widetilde{\theta}_i). \label{eqn:max_particles}\\
    &= \min_\psi \max_{w \in \R^N, \widetilde{\theta}_{1:N} \subset \Omega} \E_{i \sim \text{SOFTMAX}(w)} \left[ \ell'(\pi^\psi, \widetilde{\theta}_i) \right]. \label{eqn:multinomial_particles}
\end{align}
In the first equality we replace the maximum over all $\Omega$ to a maximum over all subsets $\widetilde{\Theta} = \widetilde{\theta}_{1:N}$ of size $N$. 
The resulting maximum over the $N$ points is still a discrete optimization. 
To smooth it out, we utilize the fact that a $\max$ over a set is just the same as the maximum over of the expectation over all distributions on that set. 
In the last equality, we reparameterize the set of distributions with the softmax to weight the different values of $\widetilde{\theta}$. 
In each round, we backpropagate through $w$ and $\widetilde{\theta}_{1:N}$.
Now we discuss the optimization routine outlined in Algorithm~\ref{alg:optim}. For the inside optimization, ideally, in each round we would build an estimate of the loss function at our current choice of $\pi^{\psi}$ for each of the $\widetilde{\theta}_{1:N}$'s under consideration. To do so, we rollout the policy for each $\theta\in \widetilde{\theta}_{1:N}$ under consideration $L$ times and then average the resulting losses (this also allows us to construct a stochastic gradient of the loss). In practice we can't consider all $\theta\in  \widetilde{\theta}_{1:N}$, so instead we sample $M$ of them from $w$. This has a computational benefit by allowing us to be strategic by considering $\theta$'s each round that are closest to the $\arg\max_{\widetilde\theta_{1:N}} \ell'(\pi^{\psi}, \theta)$.



After this we then backpropagate through $w$ and $\widetilde{\Theta}$ using the stochastic gradients learned from the rollouts. Finally, we then update $\pi$ by backpropagation through the neural network under consideration. The gradient steps are taken with unbiased gradient estimates $g^w(i, \tau)$, $g^{\widetilde{\Theta}}(i, \tau)$ and $g^\psi(i, \tau)$, which are computed by using the score-function identity and is described in detail in Appendix~\ref{sec:grad_deriv}. We outline more implementation details in Appendix \ref{sec:optim} along with the below algorithm with explicit gradient estimate formulas. Hyperparamters can be found in Appendix~\ref{sec:hyperparam}.

\begin{algorithm}
\small
\caption{Gradient Based Optimization of \eqref{eqn:multinomial_particles}}
\label{alg:optim}

\begin{algorithmic}

\STATE {\bfseries Input:} partition $\Omega$, number of iterations $N_{it}$, number of problem samples $M$, number of rollouts per problem $L$, and loss variable $L_T$ at horizon $T$ (see beginning of Section~\ref{sec:framework}).

\STATE {\bfseries Goal}: Compute the optimal policy $\arg\min_{\pi} \max_{\theta \in \Omega} \ell'(\pi, \theta) = \arg\min_{\pi} \max_{\theta \in \Omega} \mathbb{E}_{\pi, \theta}[L_T]$. Note in the case of $\ell'(\pi, \theta) = \ell(\pi, \theta) - b^{(r_{k(\theta)})}$, $L_T$ is inherently subtracting the scalar value $b^{(r_{k(\theta)})}$.

\STATE {\bfseries Initialization:} $w$, finite set $\widetilde{\Theta} = \widetilde{\theta}_{1:N}$ and $\psi$.

\FOR{$t = 1, ..., N_{\text{it}}$}
    
    \FOR{$m = 1, ..., M$}
        \STATE Sample $I_{m} \overset{i.i.d.}{\sim} \text{SOFTMAX}(w)$.\\
        \STATE Collect $L$ independent rollout trajectories, denoted as $\tau_{m, 1:L}$, by the policy $\pi^\psi$ for $\theta_{I_m}$.
    \ENDFOR

    
    \STATE Update the generating distribution by taking ascending steps on gradient estimates: \vspace{-1\intextsep}
    \STATE \begin{align*}
        \widetilde{\Theta}, w &\leftarrow \widetilde{\Theta} + \frac{1}{ML} \sum_{m=1}^M \left( \nabla_{\widetilde{\Theta}}\mc{L}_{\text{barrier}}(\widetilde{\theta}_{I_m}, \Omega) + \sum_{l=1}^Lg^{\widetilde{\Theta}}(I_m, \tau_{m,l})\right), w + \frac{1}{ML} \sum_{m=1}^M \sum_{l=1}^L g^w(I_m, \tau_{m,l})
    \end{align*}
    \STATE where $\mc{L}_{\text{barrier}}$ is a differentiable barrier loss that heavily penalizes the $\widetilde{\theta}_{I_m}$'s outside $\Omega$. 

    
    \STATE Update the policy by taking descending step on gradient estimate:
    \vspace{-1\intextsep}
    \STATE \begin{align*}
        \psi \leftarrow& \psi - \frac{1}{ML} \sum_{m=1}^M \sum_{l=1}^Lg^\psi(I_m, \tau_{m,l})
    \end{align*}
\ENDFOR
\end{algorithmic}
\end{algorithm}

\section{Experiments}

We now evaluate the approach described in the previous section for combinatorial bandits with $\mc{X} = \{\mb{e}_i : i \in [d]\}$ and $\mc{Z} \subset \{0,1\}^d$. This setting generalizes both binary active classification for arbitrary model class and active recommendation, which we evaluate by conducting experiments on two respective real datasets.
We evaluated based on two criteria: \emph{instance-dependent worst-case} and \emph{average-case}. For instance-dependent worst-case, we measure, for each $r_k$ and policy $\pi$, $\displaystyle\ell(\pi, \Theta^{(r_k)}) := \max_{\theta \in \Theta^{(r_k)}}\ell(\pi, \theta)$ and plot this value as a function of $r_k$. We note that our algorithm is designed to optimize for such a metric. For the secondary average-case metric, we instead measure, for policy $\pi$ and some collected set $\Theta$, $\frac{1}{|\Theta|} \sum_{\theta \in \Theta} \ell(\pi, \theta)$. Performances of instance-dependent worst-case metric are reported in Figures~\ref{fig:thresh_exp1},~\ref{fig:thresh_exp1_gap},~\ref{fig:thresh_exp2},~\ref{fig:twentyQ_exp2},~and~\ref{fig:jester_exp2} below while the average case performances are reported in the tables and Figure~\ref{fig:thresh_exp3}. Full scale of the figures can also be found in Appendix~\ref{sec:full_scale}.

\subsection{Algorithms}
We compare against a number of baseline active learning algorithms (see Section~\ref{sec:related_work} for a review).
\textsc{Uncertainty sampling} at time $t$ computes the empirical maximizer of $\langle z, \widehat{\theta} \rangle$ and the runner-up, and samples an index uniformly from their symmetric difference (i.e thinking of elements of $\mc{Z}$ as subsets of $[d]$); if either are not unique, an index is sampled from the region of disagreement of the winners (see Appendix~\ref{sec:usalg} for details). 
The greedy methods are represented by soft generalized binary search (\textsc{SGBS}) \citep{nowak2011geometry} which maintains a posterior distribution over $\mc{Z}$ and samples to maximize information gain. A hyperparameter $\beta \in (0,1/2)$ of SGBS determines the strength of the likelihood update. We plot or report a range of performance over $\beta \in \{.01, .03, .1, .2, .3, .4\}$.
The agnostic algorithms for classification \citep{balcan2006agnostic,hanneke2007teaching,hanneke2007bound,dasgupta2008general_DHM,huang2015efficient,jain2019newActiveLearning} or combinatorial bandits \citep{chen2014combinatorial,gabillon2016improved,chen2017nearly,cao2017disagreement,fiez2019sequential,jain2019newActiveLearning} are so conservative that given just $T=20$ samples, they are all exactly equivalent to uniform sampling and hence represented by \textsc{Uniform}.
To represent a policy based on learning to actively learn with respect to a prior, we employ the method of \citet{kveton2020differentiable}, denoted \textsc{Bayes-LAL}, with a fixed prior $\widetilde{\mc{P}}$ constructed by drawing a $z$ uniformly at random from $\mc{Z}$ and defining $\theta = 2z-1 \in [-1,1]^d$ (details in Appendix~\ref{sec:lal}). When evaluating each policy, we use the successive halving algorithm \citep{li2017hyperband,li2018massively} for optimizing our non-convex objective with randomly initialized gradient descent and restarts (details in Appendix~\ref{sec:hyperband}).

\subsection{Synthetic Dataset: Thresholds}
We begin with a very simple instance to demonstrate the instance-dependent performance achieved by our learned policy.
For $d=25$, let $\mc{X} = \{ \mb{e}_i : i \in [d] \}$, $\mc{Z} = \{ \sum_{i=1}^k \mb{e}_i  : k=0,1,\dots,d \}$, and $f(\cdot | \theta,x)$ is a Bernoulli distribution over $\{-1,1\}$ with mean $\langle x, \theta \rangle \in [-1,1]$. 
Appendix~\ref{sec:instance_dep_applications} shows that $z_{\ast}(\theta_{\ast}) = \arg\max_{z} \langle z, \theta_{\ast}\rangle$ is the best threshold classifier for a label distribution induced by $(\theta_{\ast}+1)/2$.
We trained baseline policies $\{\pi_k\}_{k=1}^9$ for the \textsc{Best Identification} metric with $\mc{C}(\theta) = \widetilde{\rho}(\mc{X}, \mc{Z}, \theta)$ and $r_k = 2^{3 + i/2}$ for $i\in\{0,\dots,8\}$.

\begin{wrapfigure}{r}{.65\textwidth}
\vspace{-2\intextsep}
\begin{minipage}[c]{.49\linewidth}
        \centering
        \includegraphics[trim={0.5cm 1cm 0.5cm 0.8cm},clip,width=\linewidth]{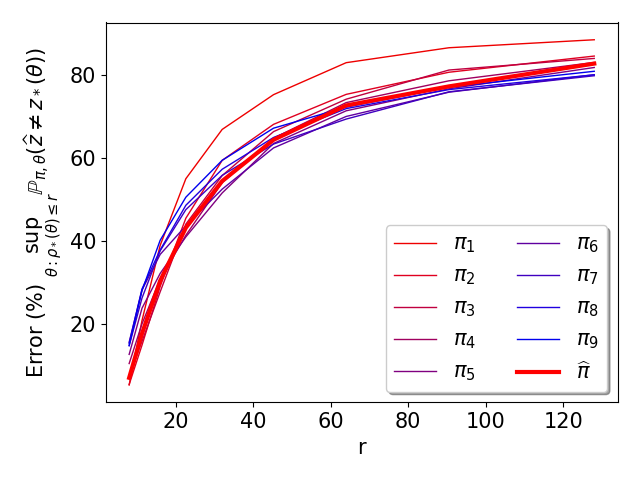}
        \captionof{figure}{Learned policies, lower is better}
        \label{fig:thresh_exp1}
\end{minipage}
\begin{minipage}[c]{.49\linewidth}
        \centering
        \includegraphics[trim={0.5cm 0cm 0.5cm 0},clip,width=\linewidth]{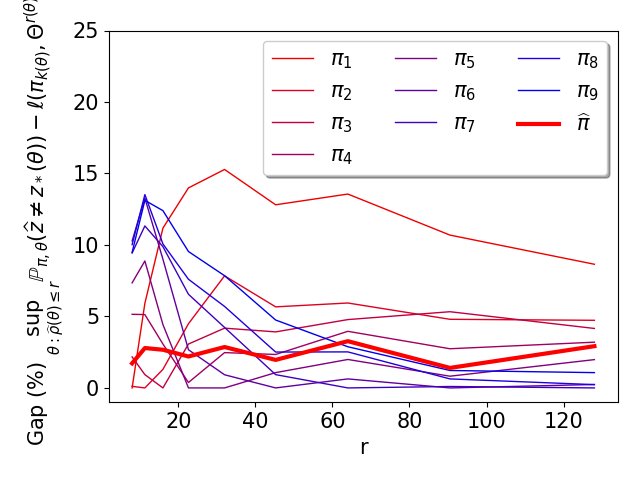}
        \captionof{figure}{Sub-optimality of individual policies, lower is better}
        \label{fig:thresh_exp1_gap}
\end{minipage}
\begin{minipage}[c]{.49\linewidth}
        \centering
        \includegraphics[trim={0.3cm 1cm 0.5cm 0},clip,width=\linewidth]{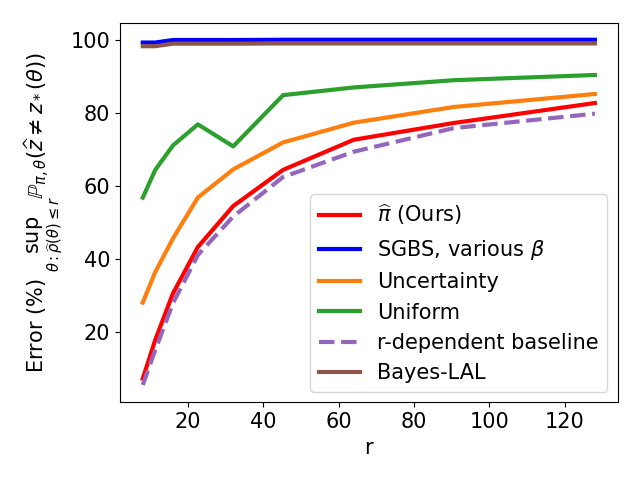}
        \captionof{figure}{Max $\{\theta:\widetilde{\rho}(\theta)\leq r \}$, lower is better}
        \label{fig:thresh_exp2}
\end{minipage}
\begin{minipage}[c]{.49\linewidth}
        \centering
        \includegraphics[trim={0.3cm 0cm 0.5cm 1cm},clip,width=\linewidth]{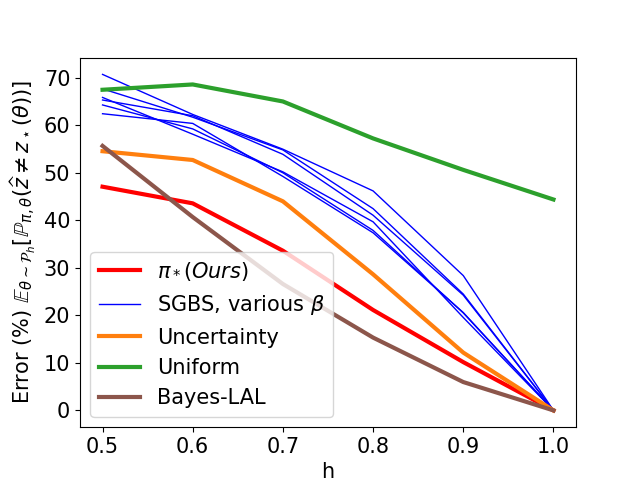}
        \captionof{figure}{Average $\E_{\theta \sim \mc{P}_h}[ \cdot ]$, lower is better}
        \label{fig:thresh_exp3}
\end{minipage}
\vspace{-3\intextsep}
\end{wrapfigure}

First we compare the base policies $\pi_k$ to $\widehat{\pi}$. 
Figure~\ref{fig:thresh_exp1} presents $\ell(\pi,\Theta^{(r)})=\max_{\theta : \widetilde{\rho}(\theta) \leq r} \ell(\pi,\theta) = \max_{\theta : \widetilde{\rho}(\theta) \leq r} \P_{\pi,\theta}( \widehat{z} \neq z_\ast(\theta) )$ as a function of $r$ for our base policies $\{ \pi_k \}_k$ and the global policy $\widehat{\pi}$, each as an individual curve. 
Figure~\ref{fig:thresh_exp1_gap} plots the same information in terms of gap: $\displaystyle \ell(\pi,\Theta^{(r)}) - \min_{k:  r_{k-1} < r \leq r_k }\ell(\pi_k,\Theta^{(r_k)})$.
We observe that each $\pi_k$ performs best in a particular region and $\widehat{\pi}$ performs almost as well as the $r$-dependent baseline policies over the range of $r$. 

Under the same conditions as Figure~\ref{fig:thresh_exp1}, Figure~\ref{fig:thresh_exp2} compares the performance of $\widehat{\pi}$ to the algorithm benchmarks.
Since SGBS and Bayes-LAL are deterministic, the adversarial training finds a $\theta$ that tricks them into catastrophic failure. 
Figure~\ref{fig:thresh_exp3} trades adversarial evaluation for evaluating with respect to a parameterized prior: For each $h \in \{0.5,0.6,\dots,1\}$, $\theta \sim \mc{P}_h$ is defined by drawing a $z$ uniformly at random from $\mc{Z}$ and then setting $\theta_i = (2z_i - 1)(2\alpha_i-1)$ where $\alpha_i \sim \text{Bernoulli}(h)$. Thus, each sign of $2z-1$ is flipped with probability $h$. We then compute $\E_{\theta \sim \mc{P}_h}[\P_{\pi,\theta}( \widehat{z} = z_\ast(\theta) )] = \E_{\theta \sim \mc{P}_h}[\ell(\pi,\theta)]$. 
While SGBS now performs much better than uniform and uncertainty sampling, our policy $\widehat{\pi}$ is still superior to these policies. However, Bayes-LAL is best overall which is expected since the support of $\mc{P}_h$ is essentially a rescaled version of the prior used in Bayes-LAL.

\subsection{Real Datasets}
\noindent \textbf{20 Questions.}
Our dataset is constructed from the real data of \citet{hu2018playing}. 
Summarizing how we used the data, $100$ yes/no questions were considered for $1000$ celebrities. Each question $i \in [100]$ for each person $j \in [1000]$ was answered by several annotators to construct an empirical probability $\bar{p}_{i}^{(j)} \in [0,1]$ denoting the proportion of annotators that answered ``yes.'' 
To construct our instance, we take $\mc{X} = \{ \mb{e}_i : i \in [100] \}$ to encode questions and $\mc{Z} = \{ z^{(j)} : [z^{(j)}]_i = \1\{ \bar{p}_{i}^{(j)} > 1/2 \} \} \subset \{0,1\}^{1000}$.
Just as before, we trained $\{\pi_k\}_{k=1}^4$ for the \textsc{Best Identification} metric with $\mc{C}(\theta) = \widetilde{\rho}(\mc{X}, \mc{Z}, \theta)$ and $r_i = 2^{3 + i/2}$ for $i\in\{1,\dots,4\}$. See Appendix~\ref{sec:20qsetup} for details.

\noindent \textbf{Jester Joke Recommendation.}
We now turn our attention away from \textsc{Best Identification} to \textsc{Simple Regret} where $\ell(\pi,\theta) = \E_{\pi,\theta}[ \langle z_\ast(\theta) - \widehat{z}, \theta \rangle ]$.
We consider the Jester jokes dataset of \citet{goldberg2001eigentaste} that contains jokes ranging from innocent puns to grossly offensive jokes. 
We filter the dataset to only contain users that rated all $100$ jokes, resulting in 14116 users.
A rating of each joke was provided on a $[-10,10]$ scale which was rescaled to $[-1,1]$ and observations were simulated as Bernoulli's like above.
We then clustered the ratings of these users (see Appendix~\ref{sec:jester_setup} for details) to 10 groups to obtain $\mc{Z} = \{z^{(k)} : k\in[10], z^{(k)} \in\{0,1\}^{100}\}$ where $z_{i}^{(k)} = 1$ corresponds to recommending the $i$th joke in user cluster $z^{(k)} \in \mc{Z}$. See Appendix~\ref{sec:jester_setup} for details.

\subsubsection{Instance-dependent Worst-case}

\begin{wrapfigure}{r}{.65\textwidth}
\begin{minipage}[l]{.49\linewidth}
    \centering
    \includegraphics[trim={0.3cm 1cm 0.2cm 0},clip,width=\linewidth]{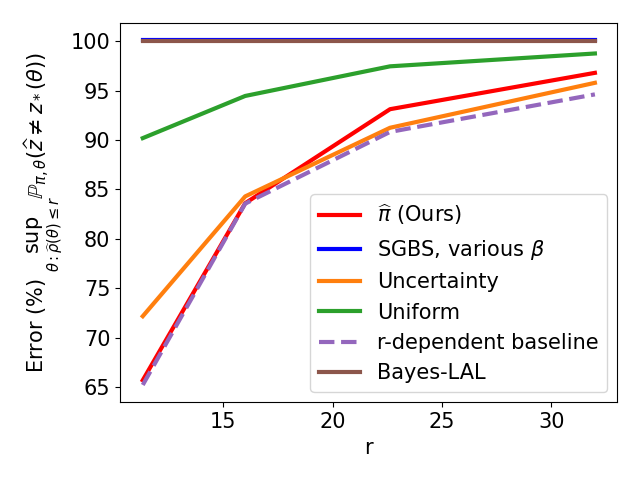}
    \captionof{figure}{20 Questions}
    \label{fig:twentyQ_exp2}
\end{minipage}
\begin{minipage}[l]{.49\linewidth}
        \centering
        \includegraphics[trim={0.3cm 0cm 1cm 2cm},clip,width=\linewidth]{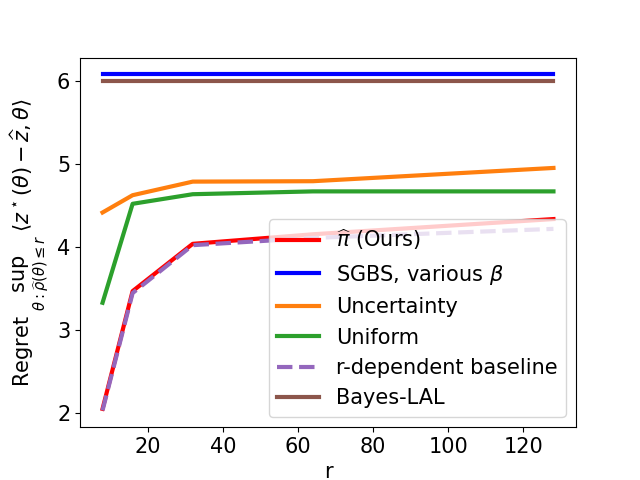}
        \captionof{figure}{Jester Joke}
        \label{fig:jester_exp2}
\end{minipage}
\vspace{-2\intextsep}
\end{wrapfigure}

Figure~\ref{fig:twentyQ_exp2} and Figure~\ref{fig:jester_exp2} are analogous to Figure~\ref{fig:thresh_exp2} but for the 20 questions and Jester joke instances, respectively. 
The two deterministic policies, SGBS and Bayes-LAL, fail on these datasets as well against the worst-case instances.
On the Jester joke dataset, our policy alone nearly achieves the $r$-dependent baseline for all $r$.
But on 20 questions, uncertainty sampling performs remarkably well. 
These experiments on real datasets demonstrate that our policy obtains near-optimal instance dependent sample complexity. 

\subsubsection{Average Case Performance}
While the metric of the previous section rewarded algorithms that perform uniformly well over all possible environments that could be encountered, in this section we consider the performance of an algorithm with respect to a distribution over environments, which we denote as average case. 

\begin{minipage}[l]{.49\linewidth}
\captionof{table}{20 Questions, higher the better}
\centering
\small
\begin{tabular}{ll}
\hline
\normalsize Method & \normalsize Accuracy(\%) \\ \hline
$\pi^*$ (Ours) &  17.9\\
SGBS& \{26.5, 26.2, 27.2,\\
& 26.5, 21.4, 12.8\}\\
Uncertainty&  14.3 \\ 
Bayes-LAL&  4.1 \\
Uniform&  6.9\\
\hline
\end{tabular}
\label{tab:twentyQ_exp4}
\end{minipage}
\begin{minipage}[l]{.49\linewidth}
\captionof{table}{Jester Joke, lower the better}
\centering
\small
\begin{tabular}{ll}
\hline
\normalsize Method & \normalsize Regret \\ \hline
$\pi^*$ (Ours) &  3.209\\
SGBS& \{3.180, 3.224, 3.278,\\
& 3.263, 3.153, 3.090\}\\
Uncertainty&  3.027 \\
Bayes-LAL&  3.610 \\
Uniform&  3.877\\
\hline
\end{tabular}
\label{tab:jester_exp4}
\end{minipage}

While heuristic based algorithms (such as SGBS, uncertainty sampling and Bayes-LAL) can perform catastrophically for worst-case instances, they can perform very well with respect to a benign distribution over instances. 
Here we demonstrate that our policy not only performs optimally under the instance-dependent worst-case metric but also remain comparable even when evaluated under the average case metric.
To measure the average performance, we construct prior distributions $\widehat{\mc{P}}$ based on the individual datasets:
\begin{itemize}
    \item For the 20 questions dataset, to draw a $\theta \sim \widehat{\mc{P}}$, we uniformly at random select a $j\in[1000]$ and sets $\theta_i = 2\bar{p}_i^{(j)}-1$ for all $i\in[d]$.
    \item For the Jester joke recommendation dataset, to draw a $\theta \sim \widehat{\mc{P}}$, we uniformly sample a user and employ their ratings to each joke.
\end{itemize}
On the 20 questions dataset, as shown in Table~\ref{tab:twentyQ_exp4}, SGBS and $\widehat{\pi}$ are the winners. 
Bayes-LAL performs much worse in this case, potentially because of the distribution shift from $\widetilde{\mc{P}}$ (prior we train on) to $\widehat{\mc{P}}$ (prior at test time).
The strong performance of SGBS may be due to the fact that $\text{sign}(\theta_i) = 2 z_\ast(\theta)_i -1$ for all $i$ and $\theta \sim \widehat{\mc{P}}$, a realizability condition under which SGBS has strong guarantees \citep{nowak2011geometry}. On the Jester joke dataset, Table~\ref{tab:jester_exp4} shows that despite our policy not being trained for this setting, its performance is still among the top.

\section{Related work}\label{sec:related_work}

\textbf{Learning to actively learn.}
Previous works vary in how the parameterize the policy, ranging from parameterized mixtures of existing expertly designed active learning algorithms \citep{baram2004online, hsu2015active, agarwal2016corralling}, parameterizing hyperparameters (e.g., learning rate, rate of forced exploration, etc.) in an existing popular algorithm (e.g, EXP3) \citep{ konyushkova2017learning, bachman2017learning,cella2020meta}, and the most ambitious, policies parameterized end-to-end like in this work \citep{boutilier2020differentiable,kveton2020differentiable, sharaf2019meta, fang2017learning, woodward2017active}. 
These works take an approach of defining a prior distribution either through past experience (meta-learning) or expert created (e.g., $\theta \sim \mc{N}(0,\Sigma)$), and then evaluate their policy with respect to this prior distribution. Defining this prior can be difficult, and moreover, if the $\theta$ encountered at test time did not follow this prior distribution, performance could suffer significantly. Our approach, on the other hand, takes an adversarial training approach and can be interpreted as learning a parameterized least favorable prior \citep{wasserman2013all}, thus gaining a much more robust policy as an end result.




\textbf{Robust and Safe Reinforcement Learning.}
Our work is also highly related to the field of robust and safe reinforcement learning, where our objective can be considered as an instance of \textit{minimax criterion under parameter uncertainty} \citep{garcia2015comprehensive}. 
Widely applied in applications such as robotics \citep{mordatch2015ensemble, rajeswaran2016epopt}, these methods train a policy in a simulator like Mujoco \citep{todorov2012mujoco} to minimize a defined loss objective while remaining robust to uncertainties and perturbations to the environment \citep{mordatch2015ensemble, rajeswaran2016epopt}.
Ranges of these uncertainty parameters are chosen based on potential values that could be encountered when deploying the robot in the real world. 
In our setting, however, defining the set of environments is far less straightforward and is overcome by the adoption of the $\mc{C}(\theta)$ function.

\textbf{Active Binary Classification Algorithms.}
The literature on active learning algorithms can be partitioned into \emph{model-based heuristics} like uncertainty sampling, query by committee, or model-change sampling \citep{settles2009active}, \emph{greedy binary-search} like algorithms that typically rely on a form of bounded noise for correctness \citep{dasgupta2005analysis,dasgupta2006coarse_splittingindex,kaariainen2006active,golovin2011adaptive,nowak2011geometry}, and \emph{agnostic} algorithms that make no assumptions on the probabilistic model \citep{balcan2006agnostic,hanneke2007teaching,hanneke2007bound,dasgupta2008general_DHM,huang2015efficient,jain2019newActiveLearning,katz2020empirical,katz2021improved}. 
Though the heuristics and greedy methods can perform very well for some problems, it is typically easy to construct counter-examples (e.g., outside the assumptions) in which they catastrophically fail as demonstrated in our experiments.
The agnostic algorithms have strong robustness guarantees but rely on concentration inequalities, and consequently require at least hundreds of labels to observe any deviation from random sampling (see \citet{huang2015efficient} for comparison). 
Therefore, they were implicitly represented by uniform in our experiments.

\textbf{Pure-exploration Multi-armed Bandit Algorithms.}
In the linear structure setting, for sets $\mc{X},\mc{Z} \subset \R^d$ known to the player, pulling an ``arm'' $x \in \mc{X}$ results in an observation $\langle x, \theta_* \rangle + $ zero-mean noise, and the objective is to identify $\arg\max_{z \in \mc{Z}} \langle z, \theta_* \rangle$ for a vector $\theta_*$ unknown to the player \citep{soare2014best,karnin2016verification,tao2018best,xu2017fully,fiez2019sequential}. 
A special case of linear bandits is combinatorial bandits where $\mc{X} = \{ \mb{e}_i : i \in [d] \}$ and $\mc{Z} \subset \{0,1\}^d$ \citep{chen2014combinatorial,gabillon2016improved,chen2017nearly,cao2017disagreement,fiez2019sequential,jain2019newActiveLearning}.
Active binary classification is a special case of combinatorial pure-exploration multi-armed bandits \citep{jain2019newActiveLearning}, which we exploit in the threshold experiments.
While the above works have made great theoretical advances in deriving algorithms and information theoretic lower bounds that match up to constants, the constants are so large that these algorithms only behave well when the number of measurements is very large. 
When applied to the instances of our paper (only 20 queries are made), these algorithms behave no differently than random sampling.

\section{Discussion and Future Directions}
We see this work as an exciting but preliminary step towards realizing the full potential of this general approach.
Although our experiments has been focusing on applications of combinatorial bandit, we see this framework  generalizing with minor changes to many more widely applicable settings such as multi-class active classification, contextual bandits, etc. To generalize $\mc{C}(\theta)$ to these settings, one can refer to existing literature for instance-dependent lower bounds \citep{katz2021improved,agarwal2014taming}.
Alternatively, when such a lower bound does not exist, we conjecture that a heuristic scoring function could also serve as $\mc{C}(\theta)$. For example, in a chess game, one could simply use the scoring function of the pieces left on board as a proxy for difficulty.

From a practical perspective, training a $\widehat{\pi}$ can take many hours of computational resources for even these small instances. 
Scaling these methods to larger instances is an important next step.
While training time scales linearly with the horizon length $T$, we note that one can take multiple samples per time step. With minimal computational overhead, this could enable training on problems that require larger sample complexities.
In our implementation we hard-coded the decision rule for $\widehat{z}$ given $s_T$, but it could also be learned as in \citep{Luedtkeeaaw2140}.
Likewise, the parameterization of the policy and generator worked well for our purposes but was chosen somewhat arbitrarily--are there more natural choices? 
Finally, while we focused on stochastic settings, this work naturally extends to constrained fully adaptive adversarial sequences which is an interesting direction of future work.

\section*{Funding disclosure}
This work was supported in part by NSF IIS-1907907.

\section*{Acknowledgement}
The authors would like to thank Aravind Rajeswaran for insightful discussions on Robust Reinforcement Learning. We would also like to thank Julian Katz-Samuels and Andrew Wagenmaker for helpful comments.


\bibliography{main}
\bibliographystyle{iclr2022_conference}

\newpage

\begin{appendix}
\section{Instance Dependent Sample Complexity}\label{sec:instance_dep_applications}

Identifying forms of $\mc{C}(\theta)$ is not as difficult a task as one might think due to the proliferation of tools for proving lower bounds for active learning \citep{mannor2004sample,tsybakov2008introduction,garivier2016optimal,carpentier2016tight,simchowitz2017simulator,chen2014combinatorial}. One can directly extract values of $\mc{C}(\theta)$ from the literature for regret minimization of linear or other structured bandits \citep{lattimore2016end,van2020optimal}, contextual bandits \citep{hao2019adaptive}, and tabular as well as structured MDPs \citep{simchowitz2019non,ok2018exploration}.
Moreover, we believe that even reasonable surrogates of $\mc{C}(\theta)$  should result in a high quality policy $\pi_*$.

We review some canonical examples:
\begin{itemize}
    \item \textbf{Multi-armed bandits.} In the best-arm identification problem, there are $d \in \mathbb{N}$ Gaussian distributions where the $i$th distribution has mean $\theta_i \in \R$ for $i=1,\dots,d$. In the above formulation, this problem is encoded as action $x_t = i_t$ results in observation $y_t \sim \text{Bernoulli}(\theta_{i_t})$ and the loss $\ell(\pi,\theta) := \E_{\pi,\theta}[ \1\{ \widehat{i} \neq i_\ast(\theta) \} ]$ where $\widehat{i}$ is $\pi$'s recommended index and $i_\ast(\theta) = \arg\max_{i} \theta_i$. 
    It's been shown that there exists a constant $c_0>0$ such that for any sufficiently large $\nu >0$ we have
    \begin{align*}
        &\min_\pi \max_{\theta : \mc{C}_{MAB}(\theta) \leq \nu}\ell(\pi,\theta) \geq \exp( - c_0 T / \nu) \\
        &\text{ where } \quad \mc{C}_{MAB}(\theta) := \sum_{i \neq i_\ast(\theta)} (\theta_{i_\ast(\theta)} - \theta_i)^{-2}
    \end{align*}
    Moreover, for any $\theta \in \R^d$ there exists a policy $\widetilde{\pi}$ that achieves $\ell(\widetilde{\pi},\theta) \leq c_1 \exp( - c_2 T / \mc{C}_{MAB}(\theta) )$ where $c_1,c_2$ capture constant and low-order terms \citep{carpentier2016tight,karnin2013almost,simchowitz2017simulator,garivier2016optimal}.
\end{itemize}
The above correspondence between the lower bound and the upper bound suggests that $\mc{C}_{MAB}(\theta)$ plays a critical role in determining the difficulty of identifying $i_\ast(\theta)$ for \emph{any} $\theta$.
This exercise extends to more structured settings as well:
\begin{itemize}
    \item \textbf{Content recommendation / active search.} Consider $n$ items (e.g., movies, proteins) where the $i$th item is represented by a feature vector $x_i \in \mc{X} \subset \R^d$ and a measurement  $x_t = x_i$ (e.g., preference rating, binding affinity to a target) is modeled as a linear response model such that  $y_t \sim \mc{N}(\langle x_i, \theta \rangle,1)$ for some unknown $\theta \in \R^d$. 
    If $\ell(\pi,\theta) := \E_{\pi,\theta}[ \1\{ \widehat{i} \neq i_\ast(\theta) \} ]$ as above then nearly identical results to that of above hold for an analogous function of $\mc{C}_{MAB}(\theta)$ \citep{soare2014best,karnin2016verification,fiez2019sequential}.
    \item \textbf{Active binary classification.} For $i=1,\dots,d$ let $\phi_i \in \R^p$ be a feature vector of an unlabeled item (e.g., image) that can be queried for its binary label $y_i \in \{-1,1\}$ where $y_i \sim \text{Bernoulli}(\theta_i)$ for some $\theta \in \R^d$. Let $\mc{H}$ be an \emph{arbitrary set of classifiers} (e.g., neural nets, random forest, etc.) such that each $h \in \mc{H}$ assigns a label $\{-1,1\}$ to each of the items $\{ \phi_i \}_{i=1}^d$ in the pool. If items are chosen sequentially to observe their labels, the objective is to identify the true risk minimizer $h_\ast(\theta) = \arg\min_{h \in \mc{H}} \sum_{i=1}^d \E_\theta[ \1\{ h(\phi_i) \neq y_i\} ]$ using as few requested labels as possible and $\ell(\pi,\theta) := \E_{\pi,\theta}[ \1\{ \widehat{h} \neq h_\ast(\theta) \} ]$ where $\widehat{h} \in \mc{H}$ is $\pi$'s recommended classifier.
    Many candidates for $\mc{C}(\theta)$ have been proposed from the agnostic active learning literature \citep{balcan2006agnostic,hanneke2007teaching,hanneke2007bound,dasgupta2008general_DHM,huang2015efficient,jain2019newActiveLearning} but we believe the most granular candidates come from the combinatorial bandit literature \citep{chen2017nearly,fiez2019sequential,cao2017disagreement,jain2019newActiveLearning}.
    To make the reduction, for each $h \in \mc{H}$ assign a $z^{(h)} \in \{0,1\}^d$ such that $[z^{(h)}]_i := \1\{ h(\phi_i) = 1 \}$ for all $i=1,\dots,d$ and set $\mc{Z} = \{ z^{(h)}: h \in \mc{H} \}$.
    It is easy to check that $z_\ast(\theta) := \arg\max_{z \in \mc{Z}} \langle z, \theta \rangle$ satisfies $z_\ast(\theta) = z^{(h_\ast(\theta))}$. Thus, requesting the label of example $i$ is equivalent to sampling from Bernoulli$(\langle \mb{e}_i, \theta \rangle) \in \{-1,1\}$, completing the reduction to combinatorial bandits: $\mc{X} = \{ \mb{e}_i : i \in [d]\}$, $\mc{Z} \subset \{0,1\}^d$.
    We then apply the exact same $\mc{C}(\theta)$ as above for linear bandits.
\end{itemize}

\section{Gradient Based Optimization Algorithm Implementation} \label{sec:optim}
First, we restate the algorithm with explicit gradient estimator formulas derived in Appendix~\ref{sec:grad_deriv}.
\begin{algorithm}
\caption{Gradient Based Optimization of \eqref{eqn:multinomial_particles} (Algorithm~\ref{alg:optim}) with explicit gradient estimators.}
\label{alg:optim2}

\begin{algorithmic}

\STATE {\bfseries Input:} partition $\Omega$, number of iterations $N_{it}$, number of problem samples $M$, number of rollouts per problem $L$, and loss variable $L_T$ at horizon $T$ (see beginning of Section~\ref{sec:framework}).

\STATE {\bfseries Goal:} Compute the optimal policy\\ $\arg\min_{\pi} \max_{\theta \in \Omega} \ell'(\pi, \theta) = \arg\min_{\pi} \max_{\theta \in \Omega} \mathbb{E}_{\pi, \theta}[L_T]$.

\STATE {\bfseries Initialization:} $w$, finite set $\widetilde{\Theta}$ and $\psi$.

\FOR{$t = 1, ..., N_{\text{it}}$}
    \STATE \textbf{Collect rollouts of play:}
    
    \FOR{$m = 1, ..., M$}
        \STATE Sample problem index $I_{m} \overset{i.i.d.}{\sim} \text{SOFTMAX}(w)$.\\
        \STATE Collect $L$ independent rollout trajectories ($\{\}$), denoted as $\tau_{m, 1:L}$, by the policy $\pi^\psi$ for problem instance $\theta_{I_m}$ and observe losses $\forall 1\leq l\leq L, L_T(\pi^{\psi}, \tau_{m, l}, \widetilde{\theta}_{I_m})$.
    \ENDFOR

    \STATE \textbf{Optimize worst cases in $\Omega$:}
    
    \STATE Update the generating distribution by taking ascending steps on gradient estimates: \vspace{-1.5\intextsep}
    \STATE \begin{align}
        w \leftarrow w + \frac{1}{ML} \sum_{m=1}^M \nabla_w & \log(\text{SOFTMAX}(w)_{I_m})
        \cdot (\sum_{l=1}^L L_T(\pi^{\psi}, \tau_{m, l}, \widetilde{\theta}_{I_m})) \label{eqn:w_gradient}  \\
        \widetilde{\Theta} \leftarrow \widetilde{\Theta} + \frac{1}{ML} \sum_{m=1}^M \sum_{l=1}^L &
        \left( \nabla_{\widetilde{\Theta}} \mc{L}_{\text{barrier}}(\widetilde{\theta}_{I_m}, \Omega) + \nabla_{\widetilde{\Theta}}L_T(\pi^{\psi}, \tau_{m, l}, \widetilde{\theta}_{I_m})\right. \nonumber\\
        &\left. + L_T(\pi^{\psi}, \tau_{m, l}, \widetilde{\theta}_{I_m}) \cdot \nabla_{\widetilde{\Theta}} \log(\P_{\pi^\psi, \widetilde{\theta}_{I_m}}(\tau_{m, l})) \right)
        \label{eqn:theta_gradient}
    \end{align}
    \STATE where $\mc{L}_{\text{barrier}}$ is a differentiable barrier loss that heavily penalizes the $\widetilde{\theta}_{I_m}$'s outside $\Omega$. 

    \STATE \textbf{Optimize policy:}
    
    \STATE Update the policy by taking descending step on gradient estimate: \vspace{-1.5\intextsep}
    \STATE \begin{align}
        \psi \leftarrow& \psi - \frac{1}{ML} \sum_{m=1}^M \sum_{l=1}^L L_T(\pi^{\psi}, \tau_{m, l}, \widetilde{\theta}_{I_m})
        \cdot \nabla_{\psi} \log(\P_{\pi^\psi, \widetilde{\theta}_{I_m}}(\tau_{m, l})) \label{eqn:policy_gradient}.
    \end{align}
\ENDFOR
\end{algorithmic}
\end{algorithm}

In the above algorithm, the gradient estimates are unbiased estimates of the true gradients with respect to $\psi$, $w$ and $\widetilde{\Theta}$ (shown in Appendix~\ref{sec:grad_deriv}). 
We choose $N$ large enough to avoid \emph{mode collapse}, and $M,L$ as large as possible to reduce variance in gradient estimates while fitting the memory constraint. We then find the appropriate large number of optimization iterations so that the variance of the gradient estimates is reduced dramatically by averaging over time.
We use Adam optimization \citep{kingma2014adam} in taking gradient updates.

Note the decomposition for $\log(\P_{\pi^\psi, \theta'}(\tau))$ in \eqref{eqn:theta_gradient} and \eqref{eqn:policy_gradient}, where rollout $\tau = \{(x_t,y_t)\}_{t=1}^T$, and
\begin{align*}
\log&(\P_{\pi^\psi, \theta'}(\{(x_t,y_t)\}_{t=1}^T)) = \log\Big(\pi^\psi(x_1) \cdot f(y_1 | \theta', s_1) \cdot \textstyle\prod_{t=2}^T \pi^\psi(s_t, x_t)\cdot f(y_t | \theta', s_t, x_t)\Big).
\end{align*}
Here $\pi^{\psi}$ and $f$ are only dependent on $\psi$ and $\widetilde{\Theta}$ respectively. During evaluation of a fixed policy $\pi$, we are interested in solving $\max_{\theta \in \Omega} \ell'(\pi,\theta)$ by gradient ascent updates like \eqref{eqn:theta_gradient}. The decoupling of $\pi^{\psi}$ and $f$ thus enables us to optimize the objective without differentiating through a policy $\pi$, which could be non-differentiable like a deterministic algorithm.

\subsection{Implementation Details}
\textbf{Training.} When training our policies for \textsc{Best Identification}, we warm start the training with optimizing \textsc{Simple Regret}. This is because a random initialized policy performs so poorly that \textsc{Best Identification} is nearly always 1, making it difficult to improve the policy. After training $\pi_{1:K}$ in MAPO (Algorithm~\ref{alg:training}), we warm start the training of $\widehat{\pi}$ with $\widehat{\pi} = \pi_{\lfloor K/2 \rfloor}$. In addition, our generating distribution parameterizations exactly follows from Section~\ref{sec:policy_opt}.

\textbf{Loss functions.} Instead of optimizing the approximated quantity from \eqref{eqn:multinomial_particles} directly, we add regularizers to the losses for both the policy and generator. 
First, we choose the $\mc{L}_{\text{barrier}}$ in \eqref{eqn:theta_gradient} to be $\lambda_{\text{barrier}} \cdot \max\{ 0, \log(\mc{C}(\mc{X},\mc{Z},\theta)) - \log(r_k) \}$, for some large constant $\lambda_{\text{barrier}}$. 
To discourage the policy from over committing to a certain action and/or the generating distribution from covering only a small subset of particles (i.e., mode collapse), we also add negative entropy penalties to both policy's output distributions and $\text{SOFTMAX}(w)$ with scaling factors $\lambda_{\text{Pol-reg}}$ and $\lambda_{\text{Gen-reg}}$.

\textbf{State representation.}
We parameterize our state space $\mc{S}$ as a flattened $|\mc{X}| \times 3$ matrix where each row represents a distinct $x \in \mc{X}$. Specifically, at time $t$ the row of $s_t$ corresponding to some $x \in \mc{X}$ records the number of times that action $x$ has been taken $\sum_{s=1}^{t-1} \mb{1}\{ x_s = x \}$, its inverse $(\sum_{s=1}^{t-1} \mb{1}\{ x_s = x \})^{-1}$, and the sum of the observations $\sum_{s=1}^{t-1} \mb{1}\{ x_s = x \} y_s$.

\textbf{Policy MLP architecture.}
Our policy $\pi^\psi$ is a multi-layer perceptron with weights $\psi$. The policy take a $3|\mc{X}|$ sized state as input and outputs a vector of size $|\mc{X}|$ which is then pushed through a soft-max to create a probability distribution over $\mc{X}$. 
At the end of the game, regardless of the policy's weights, we set $\widehat{z} = \argmax_{z \in \mc{Z}} \langle z, \widehat{\theta} \rangle$ where $\widehat{\theta}$ is the minimum $\ell_2$ norm solution to $\argmin_{\theta} \sum_{s=1}^T (y_s - \langle x_s, \theta \rangle)^2$. 

Our policy network is a simple 6-layer MLP, with layer sizes $\{3|\mc{X}|, 256, 256, 256, 256, |\mc{X}|\}$ where $3|\mc{X}|$ corresponds to the input layer and $|\mc{X}|$ is the size of the output layer, which is then pushed through a Softmax function to create a probability over arms. In addition, all intermediate layers are activated with the leaky ReLU activation units with negative slopes of $.01$. For the experiments for 1D thresholds and 20 Questions, they share the same network structure as mentioned above with $|\mc{X}| = 25$ and $|\mc{X}| = 100$ respectively.

\section{Gradient Estimate Derivation} 

\label{sec:grad_deriv}
Here we derive the unbiased gradient estimates \eqref{eqn:w_gradient}, \eqref{eqn:theta_gradient} and \eqref{eqn:policy_gradient} in Algorithm~\ref{alg:optim}. Since each the gradient estimates in the above averages over $M\cdot L$ identically distributed trajectories, it is therefore sufficient to show that our gradient estimate is unbiased for a single problem $\widetilde{\theta}_i$ and its rollout trajectory $\{(x_t,y_t)\}_{t=1}^T$.

For a feasible $w$, using the score-function identity \citep{aleksandrov1968stochastic}
\begin{align*}
    &\nabla_w \E_{i \sim \text{SOFTMAX}(w)} \left[ \ell(\pi^\psi, \widetilde{\theta}_i) \right] \\
    &= \E_{i \sim \text{SOFTMAX}(w)} \left[ \ell(\pi^\psi, \widetilde{\theta}_i) \cdot \nabla_w \log(\text{SOFTMAX}(w)_i) \right].
\end{align*}
Observe that if $i \sim \text{SOFTMAX}(w)$ and $\{(x_t,y_t)\}_{t=1}^T$ is the result of rolling out a policy $\pi^\psi$ on $\widetilde{\theta}_i$ then 
\begin{align*}
g^w(i, \{(x_t,y_t)\}_{t=1}^T) := L_T(\pi^\psi,\{(x_t,y_t)\}_{t=1}^T,\widetilde{\theta}_i) \cdot \nabla_w \log(\text{SOFTMAX}(w)_i)
\end{align*}
is an unbiased estimate of $\nabla_w \E_{i \sim \text{SOFTMAX}(w)} \left[ \ell(\pi^\psi, \widetilde{\theta}_i) \right]$.

For a feasible set $\widetilde{\Theta}$, by definition of $\ell(\pi,\theta)$,
\begin{align}
    \nabla_{\widetilde{\Theta}} \E_{i \sim \text{SOFTMAX}(w)}& \left[ \ell(\pi^\psi, \widetilde{\theta}_i) \right] \\
    = \E_{i \sim \text{SOFTMAX}(w)} &\left[ \nabla_{\widetilde{\Theta}} \E_{\pi,\widetilde{\theta}_i}\left[ L_T(\pi,\{(x_t,y_t)\}_{t=1}^T,\widetilde{\theta}_i) \right] \right]  \nonumber\\
    = \E_{i \sim \text{SOFTMAX}(w)} &\left[ \E_{\pi,\widetilde{\theta}_i}\left[ \nabla_{\widetilde{\Theta}}L_T(\pi,\{(x_t,y_t)\}_{t=1}^T,\widetilde{\theta}_i) + L_T(\pi,\{(x_t,y_t)\}_{t=1}^T,\widetilde{\theta}_i) \cdot \nabla_{\widetilde{\Theta}}\log(\P_{\pi^{\psi},\widetilde{\theta}_i}(\{(x_t,y_t)\}_{t=1}^T))\right] \right]\nonumber
\end{align}
where the last equality follows from chain rule and the score-function identity \citep{aleksandrov1968stochastic}. The quantity inside the expectations, call it $g^{\widetilde{\Theta}}(i, \{(x_t,y_t)\}_{t=1}^T)$, is then an unbiased estimator of $\nabla_{\widetilde{\Theta}} \E_{i \sim \text{SOFTMAX}(w)} \left[ \ell(\pi^\psi, \widetilde{\theta}_i) \right]$ given $i$ and $\{(x_t,y_t)\}_{t=1}^T$ are rolled out accordingly. Note that if $\mc{L}_{\text{barrier}}\neq0$, $\nabla_{\widetilde{\Theta}} \mc{L}_{\text{barrier}}(\widetilde{\theta}_{i}, \Omega)$ is clearly an unbiased gradient estimator of $\E_{i \sim \text{SOFTMAX}(w)} [ \E_{\pi,\widetilde{\theta}_i}[\mc{L}_{\text{barrier}}(\widetilde{\theta}_{i}, \Omega)]]$ given $i$ and rollout are sampled accordingly.

Likewise, for policy,
\begin{align*}
  g^\psi(i, \{(x_t,y_t)\}_{t=1}^T) := L_T(\pi^\psi,\{(x_t,y_t)\}_{t=1}^T,\widetilde{\theta}_i) \cdot \nabla_\psi \log(\P_{\pi^{\psi}, \widetilde{\theta}_i}(\{(x_t,y_t)\}_{t=1}^T))
\end{align*}
is an unbiased estimate of $\nabla_\psi \E_{i \sim \text{SOFTMAX}(w)} \left[ \ell(\pi^\psi, \widetilde{\theta}_i) \right]$. 

\section{Hyper-parameters} \label{sec:hyperparam}
In this section, we list our hyperparameters.
First we define $\lambda_{\text{binary}}$ to be a coefficient that gets multiplied to binary loses, so instead of $\1\{ z_\ast(\theta_*) \neq \widehat{z} \}$, we receive loss $\lambda_{\text{binary}} \cdot \1\{ z_\ast(\theta_*) \neq \widehat{z} \}$. We choose $\lambda_{\text{binary}}$ so that the recieved rewards are approximately at the same scale as $\textsc{Simple Regret}$.
During our experiments, all of the optimizers are Adam. All budget sizes are $T = 20$. For fairness of evaluation, during each experiment (1D thresholds or 20 Questions), all parameters below are shared for evaluating all of the policies. 
To elaborate on training strategy proposed in MAPO (Algorithm~\ref{alg:training}) more, we divide our training into four procedures, as indicated in Table~\ref{tab:it_lr}:
\begin{itemize}
    \item \textbf{Init.} The initialization procedure takes up a rather small portion of iterations primarily for the purpose of optimizing for $\mc{L}_{\text{barrier}}$ so that the particles converge into the constrained difficulty sets. In addition, during the initialization process we initialize and freeze $w = \vec{0}$, thus putting an uniform distribution over the particles. This allows us to utilize the entire set of particles without $w$ converge to only a few particles early on. To initialize $\widetilde{\Theta}$, we sample $2/3$ of the $N$ particles uniformly from $[-1, 1]^{|\mc{X}|}$ and the rest $1/3$ of the particles by sampling, for each $i \in [|\mc{Z}|]$, $\frac{N}{3 |\mc{Z}|}$ particles uniformly from $\{\theta: \argmax_{j} \langle \theta, z_j \rangle = i \}$. We initialize our policy weights by Xavier initialization with weights sampled from normal distribution and scaled by $.01$.
    \item \textbf{Regret Training, $\widetilde{\pi}_i$} Training with \textsc{Simple Regret} objective usually takes the longest among the Procedures. The primary purpose for this process is to let the policy converge to a reasonable warm start that already captures some essence of the task.
    \item \textbf{Fine-tune $\pi_i$.} Training with \textsc{Best Identification} objective run multiple times for each $\pi_i$ with their corresponding complexity set $\Theta_i$. During each run, we start with a warm started policy, and reinitialize the rest of the models by running the initialization procedure followed by optimizing the \textsc{Best Identification} objective.
    \item \textbf{Fine-tune $\widehat{\pi}$} This procedure optimizes \eqref{eqn:opt_policy}, with baselines $\min_k \ell(\pi_k, \Theta^{(r_k)})$ evaluated based on each $\pi_i$ learned from the previous procedure. Similar to fine-tuning each individual $\pi_i$, we warm start a policy $\pi_{\lfloor K/2 \rfloor}$ and reinitialize $w$ and $\Theta$ by running the initialization procedure again.
\end{itemize}

\begin{table*}
\centering

\begin{tabular}{@{}llccc@{}} \toprule
& & \multicolumn{2}{c}{Experiment} \\ \cmidrule(l){3-5}
Procedure & Hyper-parameter & \shortstack{1D Threshold \\$|\mc{X}|=25$} & \shortstack{20 Questions \\$|\mc{X}|=100$} & \shortstack{Jester Joke \\$|\mc{X}|=100$} \\ \midrule
\multirow{4}{*}{Init} & $N_{it}$ & \multicolumn{3}{c}{20000 (all)} \\
& $\psi$ learning rate & \multicolumn{3}{c}{$10^{-4}$ (all)} \\
& $\widetilde{\Theta}$ learning rate & \multicolumn{3}{c}{$10^{-3}$ (all)} \\
& $w$ learning rate & \multicolumn{3}{c}{$0$ (all)} \\
\midrule
\multirow{4}{*}{Regret Training} & $N_{it}$ & \multicolumn{3}{c}{480000 (all)} \\
& $\psi$ learning rate & \multicolumn{3}{c}{$10^{-4}$ (all)} \\
& $\widetilde{\Theta}$ learning rate & \multicolumn{3}{c}{$10^{-3}$ (all)} \\
& $w$ learning rate & \multicolumn{3}{c}{$10^{-3}$ (all)} \\
\midrule
\multirow{4}{*}{Fine-tune} & $N_{it}$ for $\widetilde{\pi}_i$ & 200000 & 0 & 200000 \\
& $N_{it}$ for $\pi_i$ & 200000 & 1500000 & N/A \\
& $N_{it}$ for $\pi_*$ & 500000 & 250000 & 500000 \\
& $\psi$ learning rate & \multicolumn{3}{c}{$10^{-4}$ (all)} \\
& $\widetilde{\Theta}$ learning rate & \multicolumn{3}{c}{$10^{-3}$ (all)} \\
& $w$ learning rate & \multicolumn{3}{c}{$10^{-3}$ (all)} \\
\midrule
\multirow{2}{*}{Adam Optimizer} & $\beta_1$ & \multicolumn{3}{c}{.9 (all)} \\
& $\beta_2$ & \multicolumn{3}{c}{.999 (all)} \\
\bottomrule
\end{tabular}
\captionof{table}{Number of Iterations and Learning Rates}
\label{tab:it_lr}
\end{table*}

\begin{table*}
\centering
\begin{tabular}{@{}llccc@{}} \toprule
& & \multicolumn{2}{c}{Experiment} \\ \cmidrule(l){3-5}
Procedure & Hyper-parameter & \shortstack{1D Threshold \\$|\mc{X}|=25$} & \shortstack{20 Questions \\$|\mc{X}|=100$}& \shortstack{Jester Joke \\$|\mc{X}|=100$} \\ \midrule
\multirow{8}{*}{\shortstack{Init + \\ Train + \\ Fine-tune}} & $N$ & $1000\times |\mc{Z}|$ & $300 \times |\mc{Z}|$ & $2000 \times |\mc{Z}|$ \\
& M & 1000 & 500 & 500 \\
& L & 10 & 30 & 30 \\
& $\lambda_{\text{binary}}$ & 7.5 & 30 & 30 \\
& $\lambda_{\text{Pol-reg}}$(regret) & .2 & .8 & .8 \\
& $\lambda_{\text{Pol-reg}}$(fine-tune) & .3 & .8 & .8 \\
& $\lambda_{\text{Gen-reg}}$ & .05 & .1 & .05 \\
& $\lambda_{\text{barrier}}$ & \multicolumn{3}{c}{$10^3$ (all)} \\
\bottomrule
\end{tabular}
\captionof{table}{Parallel Sizes and Regularization coefficients}
\end{table*}

To provide a general strategy of choosing hyper-parameters, we note that $L$, firstly, $\lambda_{\text{binary}}$, $\lambda_{\text{Pol-reg}}$ are primarily parameters tuned for $|\mc{X}|$ as the noisiness and scale of the gradients, and entropy over the arms $\mc{X}$ grows with the size $|\mc{X}|$. Secondly, $\lambda_{\text{Gen-reg}}$ is primarily tuned for $|\mc{Z}|$ as it penalizes the entropy over the $N$ arms, which is a multiple of $|\mc{Z}|$. Thirdly, learning rate of $\theta$ is primarily tuned for the convergence of constraint $\rho^*$ into the restricted class, thus $\mc{L}_{\text{barrier}}$ becoming $0$ after the specified number of iterations during initialization is a good indicator. Finally, we choose $N$ and $M$ by memory constraint of our GPU. The hyper-parameters for each experiment was tuned with less than 20 hyper-parameter assignments, some metrics to look at while tuning these hyper-parameters includes but are not limited to: gradient magnitudes of each component, convergence of each loss and entropy losses for each regularization term (how close it is to the entropy of a uniform probability), etc.

\section{Policy Evaluation} \label{sec:hyperband}
When evaluating a policy, we are essentially solving the following objective for a fixed policy $\pi$:
\begin{align*}
\max_{\theta \in \Omega} \ell(\pi, \theta)
\end{align*}
where $\Omega$ is a set of problems. However, due to non-concavity of this loss function, gradient descent initialized randomly may converge to a local maxima. 
To reduce this possibility, we randomly initialize many initial iterates and take gradient steps round-robin, eliminating poorly performing trajectories. 
To do this with a fixed amount of computational resource, we apply the successive halving algorithm from \citet{li2018massively}. Specifically, we choose hyperparamters: $\eta = 4$, $r=100$, $R=1600$ and $s=0$. This translates to:
\begin{itemize}
    \item Initialize $|\widetilde{\Theta}|=1600$, optimize for 100 iterations for each $\widetilde{\theta}_i \in \widetilde{\Theta}$
    \item Take the top $400$ of them and optimize for another $400$ iterations
    \item Take the top $100$ of the remaining $400$ and optimize for an additional $1600$ iterations
\end{itemize}

We take gradient steps with the Adam optimizer \citep{kingma2014adam} with learning rate of $10^{-3}$ $\beta_1 = .9$ and $\beta_2 = .999$.

\section{Figures at Full Scale} \label{sec:full_scale}

\begin{minipage}[c]{.8\linewidth}
        \centering
        \includegraphics[trim={0 0.5cm 0 0},clip,width=\linewidth]{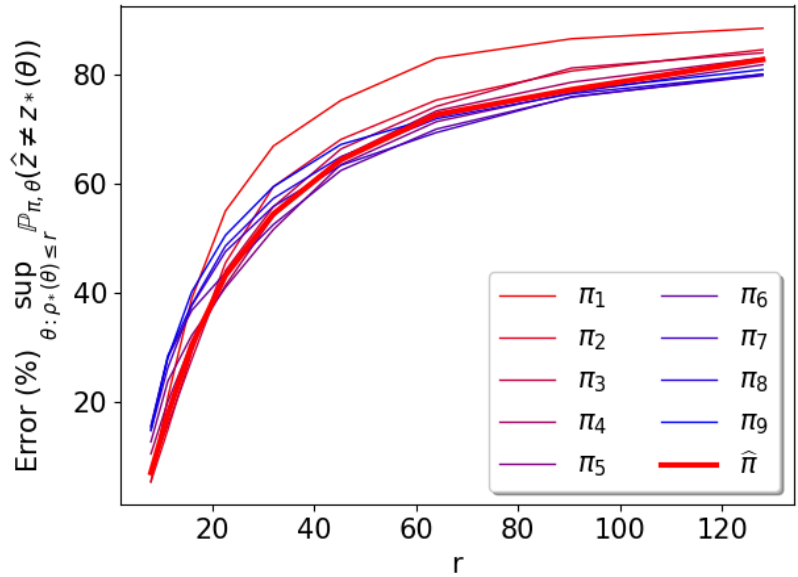}
        \captionof{figure}{Full scale of Figure~\ref{fig:thresh_exp1}}
\end{minipage}
\begin{minipage}[c]{.8\linewidth}
        \centering
        \includegraphics[trim={0 0.5cm 0 0},clip,width=\linewidth]{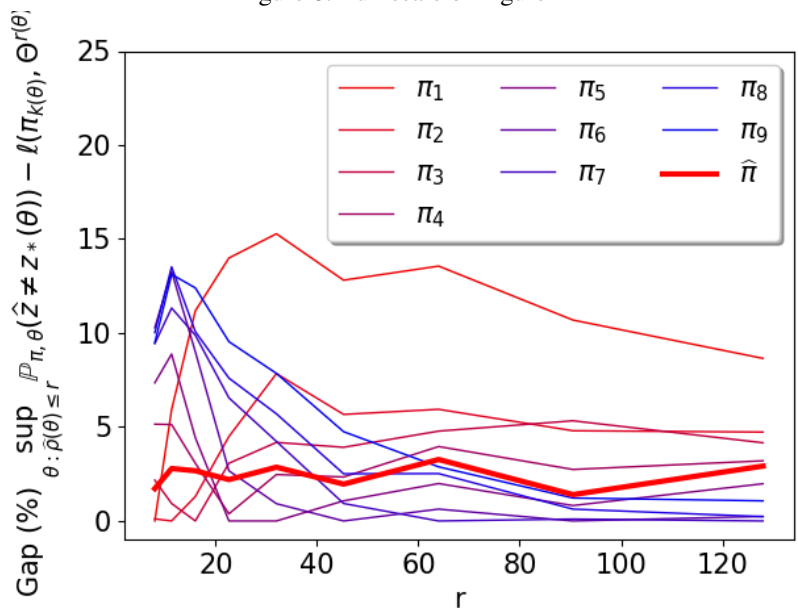}
        \captionof{figure}{Full scale of Figure~\ref{fig:thresh_exp1_gap}}
\end{minipage}
\begin{minipage}[c]{.8\linewidth}
        \centering
        \includegraphics[trim={0 1cm 0 0},clip,width=\linewidth]{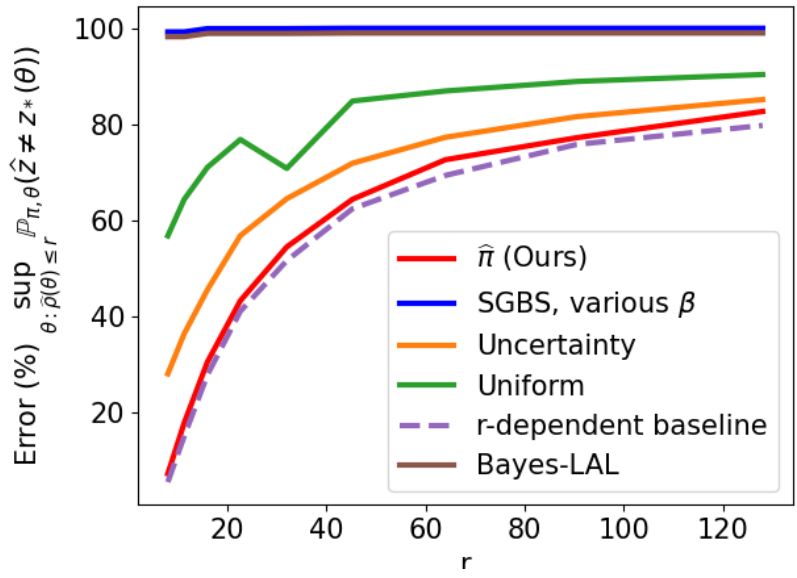}
        \captionof{figure}{Full scale of Figure~\ref{fig:thresh_exp2}}
\end{minipage}
\begin{minipage}[c]{.8\linewidth}
        \centering
        \includegraphics[trim={0 1cm 0 0},clip,width=\linewidth]{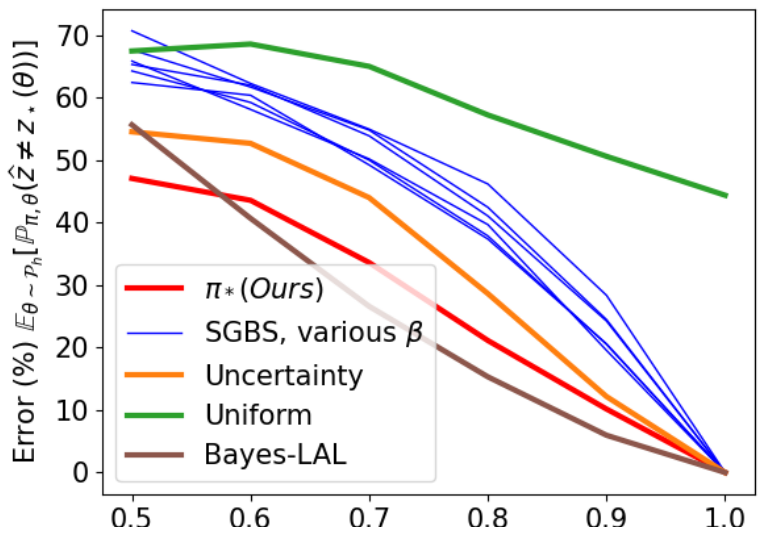}
        \captionof{figure}{Full scale of Figure~\ref{fig:thresh_exp3}}
\end{minipage}
\begin{minipage}[c]{.8\linewidth}
        \centering
        \includegraphics[trim={0 1cm 0 0},clip,width=\linewidth]{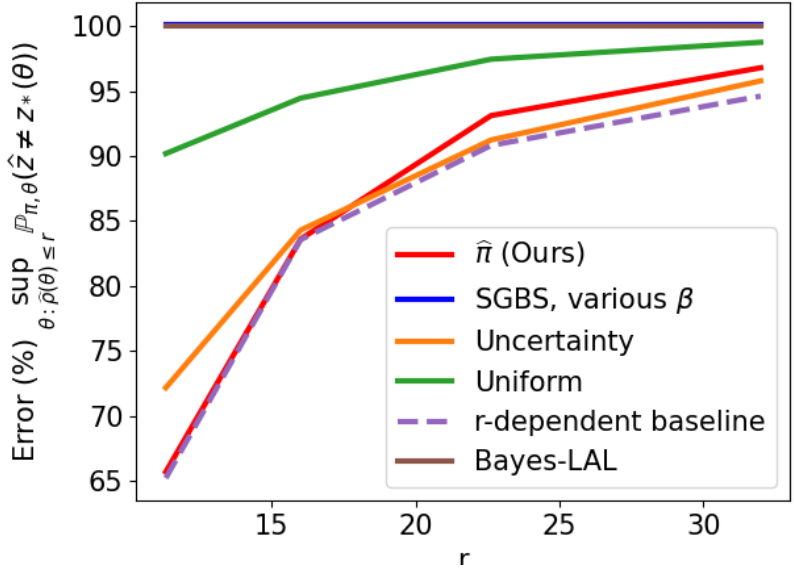}
        \captionof{figure}{Full scale of Figure~\ref{fig:twentyQ_exp2}}
\end{minipage}
\begin{minipage}[c]{.8\linewidth}
        \centering
        \includegraphics[trim={0 1cm 0 0},clip,width=\linewidth]{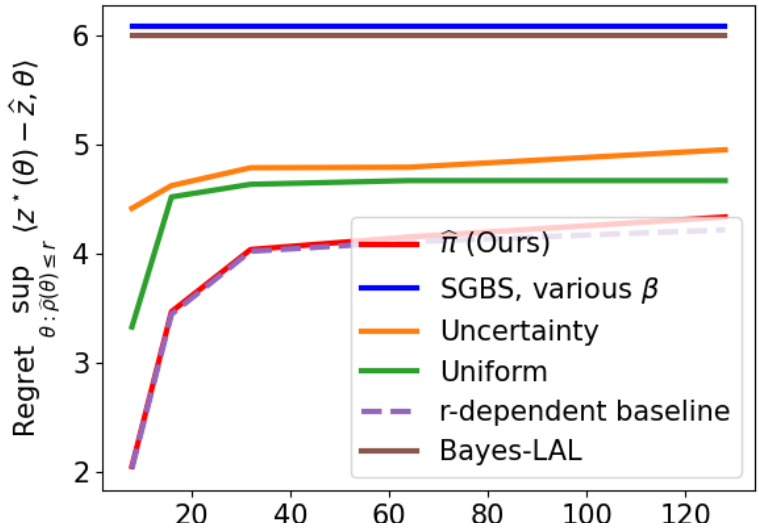}
        \captionof{figure}{Full scale of Figure~\ref{fig:jester_exp2}}
\end{minipage}

\section{Uncertainty Sampling} \label{sec:usalg}
We define the symmetric difference of a set of binary vectors, $\text{SymDiff}(\{z_1, ..., z_n\}) = \{i : \exists j,k\in[n] \ \  s.t., z_j^{(i)} = 1 \land z_k^{(i)} = 0\}$, as the dimensions where inconsistencies exist.

\begin{algorithm}[H]

\begin{algorithmic}

\STATE {\bfseries Input:} $\mc{X}, \mc{Z}$

\FOR{$t = 1, ..., T$}
    \STATE $\widehat{\theta}_{t-1} = \argmin_{\theta} \sum_{s=1}^T (y_s - \langle x_s, \theta \rangle)^2$
    
    \STATE $\widehat{\mc{Z}} = \{z \in \mc{Z}: \max_{z' \in \mc{Z}}\langle z', \widehat{\theta}_{t-1}\rangle = \langle z, \widehat{\theta}_{t-1}\rangle\}$
    
    \IF{$|\widehat{\mc{Z}}| = 1$}
    \STATE $\widehat{\mc{Z}}_t = \widehat{\mc{Z}} \bigcup \{z \in \mc{Z}: \max_{z' \in (\mc{Z}\backslash \widehat{\mc{Z}})}\langle z', \widehat{\theta}_{t-1}\rangle = \langle z, \widehat{\theta}_{t-1}\rangle\}$
    \ELSE
    \STATE $\widehat{\mc{Z}}_t = \widehat{\mc{Z}}$
    \ENDIF
    \STATE Uniformly sample $I_t$ from $\text{SymDiff}(\widehat{\mc{Z}}_t)$
    
    \STATE Pull $x_{I_t}$ and observe $y_t$
\ENDFOR
\caption{Uncertainty sampling in very small budget setting}
\label{alg:us}

\end{algorithmic}
\end{algorithm}

\section{Learning to Actively Learn Algorithm} \label{sec:lal}
To train a policy under the learning to actively learn setting, we aim to solve for the objective
\begin{align*}
\min_{\psi} \E_{\theta \sim \widehat{\mc{P}}}[\ell(\pi^\psi, \theta)]
\end{align*}
where our policy and states are parameterized the same way as Appendix~\ref{sec:representation_parameterization} for a fair comparison. To optimize for the parameter, we take gradient steps like \eqref{eqn:policy_gradient} but with the new sampling and rollout where $\widetilde{\theta}_{i} \sim \widehat{\mc{P}}$. This gradient step follows from both the classical policy gradient algorithm in reinforcement learning as well as from recent learning to actively learn work by \citet{kveton2020differentiable}.

Moreover, note that the optimal policy for the objective must be deterministic as justified by deterministic policies being optimal for MDPs. Therefore, it is clear that, under our experiment setting, the deterministic Bayes-LAL policy will perform poorly in the adversarial setting (for the same reason why SGBS performs poorly).

\section{20 Questions Setup} \label{sec:20qsetup}
\citet{hu2018playing} collected a dataset of 1000 celebrities and 500 possible questions to ask about each celebrity. We chose $100$ questions out of the $500$ by first constructing $\bar{p}'$, $\mc{X}'$ and $\mc{Z}'$ for the $500$ dimensions data, and sampling without replacement $100$ of the $500$ dimensions from a distribution derived from a static allocation. 
We down-sampled the number of questions so our training can run with sufficient $M$ and $L$ to de-noise the gradients while being prototyped with a single GPU.

Specifically, the dataset from \citet{hu2018playing} consists of probabilities of people answering \emph{Yes / No / Unknown} to each celebrity-question pair collected from some population. To better fit the combinatorial bandit scenario, we re-normalize the probability of getting \emph{Yes / No}, conditioning on the event that these people did not answer \emph{Unknown}. The probability of answering $\emph{Yes}$ to all 500 questions for each celebrity then constitutes vectors $\bar{p}'^{(1)}, ..., \bar{p}'^{(1000)} \in \R^{500}$, where each dimension of  a give $\bar{p}_i'^{(j)}$ represents the probability of yes to the $i$th question about the $j$th person. The action set $\mc{X}'$ is then constructed as $\mc{X}' = \{ \mb{e}_i : i \in [500] \}$, while $\mc{Z}' = \{ z^{(j)} : [z^{(j)}_i] = \1\{ \bar{p}_{i}^{(j)} > 1/2 \} \} \subset \{0,1\}^{1000}$ are binary vectors taking the majority votes. 

To sub-sample 100 questions from the 500, we could have uniformly at random selected the questions, but many of these questions are not very discriminative. 
Thus, we chose a ``good'' set of queries based on the design recommended by $\rho_\ast$ of \citet{fiez2019sequential}.
If questions were being answered noiselessly in response to a particular $z \in \mc{Z}'$, then equivalently we have that for this setting $\theta = 2z -1$.
Since $\rho_\ast$ optimizes allocations $\lambda$ over $\mc{X}'$ that would reduce the number of required queries as much as possible (according to the information theoretic bound of \citep{fiez2019sequential}) if we want to find a single allocation for all $z' \in \mc{Z}$ simultaneously, we can perform the optimization problem
\begin{align*}
    \min_{\lambda \in \Delta^{(|X| - 1)}} \max_{z' \in \mc{Z}'} \max_{z\neq z'}\frac{\lVert z' - z\rVert_{(\sum_i \lambda_i x_i x_i^T)^{-1}}^2}{((z' - z)^T(2z' -1))^2}.
\end{align*}
We then sample elements from $\mc{X}'$ according to this optimal $\lambda$ without replacement and add them to $\mc{X}$ until $|\mc{X}|=100$.

\section{Jester Joke Recommendation Setup} \label{sec:jester_setup}
We consider the Jester jokes dataset of \citet{goldberg2001eigentaste} that contains jokes ranging from pun-based jokes to grossly offensive. 
We filter the dataset to only contain users that rated all $100$ jokes, resulting in 14116 users.
A rating of each joke was provided on a $[-10,10]$ scale which was shrunk to $[-1,1]$.
Denote this set of ratings as $\hat{\Theta} = \{\theta_i : i\in[14116], \theta_i\in [-1, 1]^{100}\}$, where $\theta_i$ encodes the ratings of all $100$ jokes by user $i$.
To construct the set of arms $\mc{Z}$, we then clustered the ratings of these users to 10 groups to obtain $\mc{Z} = \{z_i : i\in[10], z_i \in\{0,1\}^{100}\}$ by minimizing the following metric:
\begin{align*}
\min_{\mc{Z} : |\mc{Z}| = 10} \sum_{i=1}^{14116} \max_{z_* \in \{0,1\}^{100}} \langle z_*, \theta_i \rangle - \max_{z \in \mc{Z}} \langle z, \theta_i \rangle.
\end{align*}
To solve for $\mc{Z}$, we adapt the $k-means$ algorithm, with the metric above instead of the $L-2$ metric used traditionally. 

\end{appendix}

\end{document}